Comparing Human and AI Rater Effects Using the Many-Facet Rasch Model

Hong Jiao
University of Maryland, College Park
Dan Song
Won-Chan Lee
University of Iowa

**Abstract**

Large language models (LLMs) have been widely explored for automated scoring in low-stakes assessment to facilitate learning and instruction. Empirical evidence related to which LLM produces the most reliable scores and induces least rater effects needs to be collected before the use of LLMs for automated scoring in practice. This study compared ten LLMs (ChatGPT 3.5, ChatGPT 4, ChatGPT 4o, OpenAI o1, Claude 3.5 Sonnet, Gemini 1.5, Gemini 1.5 Pro, Gemini 2.0, as well as DeepSeek V3, and DeepSeek R1) with human expert raters in scoring two types of writing tasks. The accuracy of the holistic and analytic scores from LLMs compared with human raters was evaluated in terms of Quadratic Weighted Kappa. Intra-rater consistency across prompts was compared in terms of Cronbach Alpha. Rater effects of LLMs were evaluated and compared with human raters using the Many-Facet Rasch model. The results in general supported the use of ChatGPT 4o, Gemini 1.5 Pro, and Claude 3.5 Sonnet with high scoring accuracy, better rater reliability, and less rater effects.

# Introduction

Performance assessment tasks, including essays and other types of constructed-response (CR) items are commonly used for assessing high-order thinking skills such as synthesizing information from multiple sources, critical evaluations, and evidence-based reasoning. These assessment formats are commonly employed in formative classroom assessments to facilitate instruction and learning as well as in high-stakes summative assessments for accountability and admission decisions. While the score represents a key form of feedback, the scoring of such assessment tasks is often costly, time-consuming, and error prone due to the subjectivity of human raters.

Automated scoring of performance items is an efficient and scientifically validated technology designed to complement and enhance human scoring (Shermis, 2015). For example, automating the scoring of the National Assessment of Educational Progress (NAEP) CR items could yield significant cost saving of 70% for reading and 40% for math (Justice & Palermo, 2020). As one of the most successful applications of AI technology in educational assessments, the progress in automated scoring is aligned closely with the advances in machine learning and natural language processing (NLP). Though research on automated scoring started with the pioneering work of Page (1966; 2003), recent advances in AI technology have made breakthrough in automated scoring.

Automated scoring can be categorized into two main approaches: feature-based and deep learning-based (Haller et al., 2022), with the latter further classified into neural network, small language model (SLM), and large language model (LLM) based approaches. Feature-based approaches rely on the extracted length, prompt, readability, syntactic, semantic, lexical, and discourse features (Ke & Ng, 2019; Uto et al., 2020) and supervised machine learning models. Deep learning-based approaches use word embeddings (Word2Vec, GloVe) and context embeddings from small language models (SLMs) such as BERT, deBERTa based on transformer (Vaswani et al., 2017) and large language models (LLMs) such as GPT, Claude, and Gemini.

Developing feature-based or SLM-based automated scoring systems often requires foundational skills in supervised machine learning, NLP, and deep learning. This dramatically reduces the accessibility of automated scoring to classroom learners, teachers, and assessment developers. However, with the swift emerging of more capable LLMs, automated scoring techniques are becoming increasingly accessible to non-technical users who can fully leverage the LLM technology to revamp their routine scoring practices, with enhanced efficiency and accuracy.

With the advances in Generative AI or Chatbots, LLMs have been explored for automated scoring in education assessments. LLMs have demonstrated unprecedented potential in automated scoring of essays and CR items in educational assessments (Latif & Zhai, 2024; Lee et al., 2024; Yancey et al., 2023; Yang et al., 2025). In low-stakes assessment programs, three commonly used models are ChatGPT by OpenAI, Claude by Anthropic, and Google Gemini. Recent studies have explored the use of GPT models, such as GPT 3.5 (Bui & Barrot, 2024; Mizumoto & Eguchi, 2023; Parker et al., 2023; Wijekumar et al., 2024) and GPT 4 (Kortemeyer, 2024) in automated scoring. Furthermore, several studies have compared the performance of GPT 3.5 and GPT 4 (Kim & Jo, 2024; Tate et al., 2024) as well as GPT and Llama3 (Kundu & Barbosa, 2024). Pack et al. (2024) compared PaLM 2, Claude 2, and GPT 3.5 and GPT 4, concluding that GPT 4 outperformed the others. Almost all studies on using LLMs for automated scoring purely focused on scoring accuracy, little is known about the potential rater effects of these LLM scoring engines.

Rater effects (Myford & Wolfe, 2003, 2004; Saal, Downey, & Lahey, 1980) refer to different types of inconsistencies in ratings when human raters assign scores to test-takers' constructed responses to performance tasks such as essays or CR items in educational assessments. Rater effects may occur due to raters' personal backgrounds, preferences, inaccurate interpretations of the scoring rubrics, inappropriate conceptualization of the target test-takers' population, and their idiosyncratic traits that may impact their accuracy and consistency in assigning scores to assessment items. Rater effects can be classified into different categories. Some common rater effects include rater severity/leniency and centrality. When a rater assigns scores to students' responses to a performance task consistently lower than other raters, this rater effect is noted as rater severity. When rater severity is present, students' true scores may be underestimated if such rater effect is ignored. On the other hand, when a rater assigns scores to students' responses consistently higher than other raters, this rater effect is noted as rater leniency. When rater leniency is present, students' true scores may be overestimated if such rater effect is not accounted for. Further, some raters may consistently assign scores in the middle of the score scale and not assign high scores or low scores regardless of the ability levels. This central tendency or centrality effect will introduce score errors for students whose true scores lie at the two ends of the score scale. Such rater effects are as severe as score accuracy as they may compromise the fairness, reliability, and validity of the assigned scores. Though in operational rating sessions, different strategies would be implemented to minimize rater effects such as standardized rubrics, comprehensive rater training and calibration, in-session and post-session quality control of ratings, or assigning multiple raters to each rating tasks, rater effects need to be tracked as a source of validity evidence.

Each LLM has its own strengths and features that may cater to different scoring needs and applications. Like human raters, AI raters may bear different degrees of rater effects. Thus, it is worthwhile to compare the ratings from different scoring engines and understand the potential AI rater effects of each LLM. Thus, this study aims to investigate the AI rater effects in addition to rating accuracy and consistency compared with human raters of essay scoring. The AI raters investigated in this study include ten commonly used LLMs from GPT, Gemini, and Claude families and some recent LLMs such as DeepSeek. The research questions for the study include:

1. How accurate are LLM raters compared with human raters?
2. How consistent or reliable are LLM raters compared with human raters?
3. What are the LLM rater effects compared with human raters?

## Methods

This study conducts a series of psychometric analyses to investigate the AI raters' scoring accuracy and consistency, as well as uses the Many-Facet Rasch measurement model (Linacre, 1989, 1994, 2024; Wright & Masters, 1982; Wright & Stone, 1979) to investigate and compare rater effects between AI raters and human raters in a large-scale writing assessment.

**Data**

A dataset consists of 30 student essays on two types of writing tasks. Each essay was scored by two human raters and ten AI raters, including ChatGPT 3.5, ChatGPT 4, ChatGPT 4o, OpenAI o1, Claude 3.5 Sonnet, Gemini 1.5, Gemini 1.5 Pro, Gemini 2.0, as well as DeepSeek V3 and DeepSeek R1. The 30 U.S. college students, enrolled in Mandarin Chinese language courses, included 18 third-year and 12 fourth-year students from a range of majors such as business, humanities, STEM, and diplomacy. Data collection took place over two consecutive days using constructed-response writing prompts from the 2021 and 2022 AP Chinese exams developed by the College Board. On the first day, students completed the 2021 prompts, which included one story narration (SN1) task and one email response (ER1) task. On the second day, they responded to the 2022 prompts, which featured the same two task types, each labeled as SN2, and ER2. Two certified human raters of the AP Chinese exam evaluated the writing samples using the College Board's scoring rubrics for both holistic and analytic scoring for three trait scores assessing Task Completion, Delivery, and Language Use. Each student response received both holistic and analytic trait/domain-specific scores ranging from 0 to 6 (Song & Tang, 2025).

**LLMs Trained for Automated Scoring**

In alignment with the development of LLMs, ChatGPT 3.5 was first used to score all 120 essays and other AI engines were sequentially explored when new versions or new engines were released. Data collection took approximately one and a half years. This study incorporated ten AI raters: ChatGPT 3.5, ChatGPT 4, ChatGPT 4o, OpenAI o1, Claude 3.5 Sonnet, Gemini 1.5, Gemini 2.0, Gemini Advanced 1.5 Pro, DeepSeek V3, and DeepSeek R1. These LLMs are selected based on their popularity and demonstrated expertise in processing text data and reasoning capabilities. ChatGPT 3.5 is for general-purpose tasks, casual writing, and basic coding. ChatGPT 4 is better at technical problem-solving and creative content. ChatGPT 4 generally outperforms ChatGPT 3.5 in language understanding, with ChatGPT 4 Turbo offering enhanced context handling. ChatGPT 4o excels at conversations. OpenAI o1 specializes in advanced STEM applications and debugging. Claude 3.5 Sonnet is competitive, often cited in safety-focused evaluations, technical coding, and natural-sounding content. Gemini 1.5 and 1.5 Pro are good at long-context tasks. Gemini 2.0 was selected to reflect the latest advancements in Google's LLMs, which feature enhanced language understanding and improved reasoning for research, education, and multimedia analysis. DeepSeek V3 and DeepSeek R1 are two distinct models developed by DeepSeek, each designed for different purposes and with unique capabilities. DeepSeek V3 uses the Mixture-of-Experts model, optimized for general-purpose tasks, including natural language understanding, text generation, and conversational AI. DeepSeek R1 excels in reasoning. It is specifically tailored for tasks that require deep reasoning, complex problem-solving, and advanced analytical capabilities. DeepSeek was included due to its rising prominence as an open-source LLM with demonstrated strengths in structured reasoning, grammatical precision, and technical problem-solving. Given its strong performance in formal writing tasks and its potential as a cost-effective alternative to proprietary models, incorporating DeepSeek into our analysis allows for a more comprehensive comparison of AI raters' capabilities across diverse linguistic and evaluative contexts (Gao et al., 2025).

Before scoring, human rater R1, a certified AP Chinese Language and Culture Exam rater, trained 10 AI engines (ChatGPT 3.5, ChatGPT 4.0, ChatGPT 4o, Gemini 1.5, Claude 3.5 Sonnet, OpenAI O1, Gemini 2.0, Gemini Advanced 1.5 Pro, DeepSeek V3, and DeepSeek R1) using AI training protocols designed for evaluating student samples (see Appendix A for an example protocol). As Lee et al. (2024) emphasized, human expert ratings are essential for generating benchmark data that align AI scoring with educational standards. To ensure valid AI rater performance, human rater R1 not only provided training examples drawn from the College Board website and previously scored samples but also offered guidance on applying a holistic scoring approach. AI raters were instructed to prioritize three traits: Task Completion, followed by Delivery and Language Use, in line with official rubrics. In total, 2 human raters and 10 AI raters scored 120 essays across four writing tasks.

**Analysis**

Using the analytic and holistic scores for the essays, this study investigates scoring accuracy, reliability, and rater effects. Quadratic Weighted Kappa (QWK; Cohen, 1960; 1968) was computed to evaluate the scoring accuracy of AI raters against the gold standards of human ratings. QWK was computed for each AI against each human rater. Further, AI raters' QWK were compared with the QWK measures from two human raters which were used as the baseline for comparison. Multiple ensemble models were explored by removing the worst performing LLMs from the ensemble. The first ensemble model took the average of the scores assigned by each of the studied 10 AI raters and labeled as AI11 model. Another ensemble model, AI12, was built upon AI11 model by removing model AI7, Gemini 2.0. Ensemble model, AI13, was built upon AI12 model by further removing model AI9, Gemini 1.5 Pro. Ensemble model, AI14, was built upon AI13 model by further removing model AI6, OpenAI o1. Ensemble model, AI15, was built upon AI14 model by further removing model AI8, DeepSeek V3. Ensemble model, AI16, was built upon AI15 model by further removing models AI2 and AI4, ChatGPT 4.0 and Gemini 1.5. In addition, Cronbach Alpha was computed for each AI and human ratings across the three trait scores and holistic scores as a measure of intra-rater consistency.

Furthermore, the rater effects for both human and AI raters were examined using the Many FACET Rasch model (Linacre, 1989; 1994; 2024). The FACET model for the rating scale model (Andrich, 1978) is given in equation 1.

$$\ln\left(\frac{P_{ji(x=k)}}{P_{ji(x=k-1)}}\right) = \theta_j - \tau_i - \delta_n - \beta_k. \quad (1)$$

where $\theta_j$ is student *j*'s latent ability estimate, $\tau_i$ is rater *i*'s rater effect estimate, $\delta_n$ is essay *n*'s difficulty level, $\beta_k$ is the threshold parameter for score category *k*, where the probability for a rating in category *k* is equal to the probability for a rating in category *k*-1. Our study uses the rating scale version of the FACET model (Eckes, 2015; Myford & Wolfe, 2003, 2004), assuming the threshold parameters are the same across all items. A partial credit model version can be explored by assuming that each item has its own threshold parameters. The facets investigated in this study include students (30), essays (4: SN1, ER1, SN2, ER2), raters including two human raters (R1, R2) and ten LLM-based AI raters (A1 - ChatGPT 3.5, A2 - ChatGPT 4, A3 - ChatGPT 4o, A4 - Gemini 1.5, A5 - Claude 3.5 Sonnet, A6 - OpenAI o1, A7 - Gemini 2.0, A8 - DeepSeek V3, A9 - Gemini 1.5 Pro, A10 - DeepSeek R1).

Rater effects such as the leniency or severity were quantified by the rater parameter in the FACET Rasch model. The Wright map were used to illustrate the relative difficulty of essay traits, rater effects, and student ability on the common scale. Two fit indexes were examined including the infit and outfit mean square. Infit mean square indicates the consistency a rater uses the rating scale across students and criteria while outfit mean squares quantifies the outlier score patterns. The expected values of the infit and outfit mean squares are 1. Values smaller than 1 show less variation than expected in assigned scores while values larger than 1 show more variation than expected in assigned scores. Values ranging from 0.5 to 1.5 (Linacre, 2003; Eckes, 2009) usually indicates good fit. Usually values smaller than 0.5 indicates overfitting and raters do not discriminate high performing and low performing essays while values larger than 1.5 may distort the measurement. Overfitting suggests a central tendency (Engelhard, 2002; Myford & Wolfe, 2004). Researchers also suggested more stringent criteria using 0.7 and 1.3 as the cut values (Bond & Fox, 2007; Bond et al., 2021; McNamara, 1996; Muller, 2020; Wright & Linacre, 1994). For high-stakes decision making, Myford and Wolfe (2003) even suggested that more stringent cut values like 0.8 and 1.2 might be set. This study follows the stringent criteria of 0.7 and 1.3.

## Results

**Scoring Accuracy**

The scoring accuracy of the studied AI raters was evaluated against each human rater in terms of QWK. Table 1 summarizes the holistic score accuracy in terms of QWK of each AI rater against each human rater. Two human raters' QWK was used as the baseline for comparison. The red-colored numbers indicate the highest QWK against each human rater while the blue-colored numbers indicate the second highest QWK. In general, the two human raters' QWK on essay ER2 met the criterion suggested by Williamson, Xi, and Breyer (2012); that is, QWK should be equal to or larger than a cut score of 0.7. For SN1, ChatGPT 4o yielded the highest QWK with rater 2 while Claude 3.5 yielded the highest QWK with rater 1. For SN2, Gemini 1.5 Pro yielded the highest QWK with both human raters. For ER 1, ChatGPT 4o yielded the highest QWK with rater 1 while Claude 3.5 yielded the highest QWK with rater 2. For ER2, ChatGPT 4o yielded the highest QWK with both human raters. Only one ensemble model by removing Gemini 2.0 yielded higher QWK for SN2 and ER1 with rater R1 only. In general, one of the AI rater achieved higher QWK with one human rater than human-human QWK on every item.

Table 1
Holistic Scoring Accuracy (QWK) for AI Raters against Each Human Rater

| Essay | SN1 | | SN2 | | ER1 | | ER2 | |
|---|---|---|---|---|---|---|---|---|
| Rater | R1 | R2 | R1 | R2 | R1 | R2 | R1 | R2 |
| R1/R2 | 0.683 | | 0.565 | | 0.684 | | 0.874 | |
| A1 – ChatGPT 3.5 | 0.549 | 0.409 | 0.610 | 0.538 | 0.442 | 0.516 | 0.632 | 0.560 |
| A2 – ChatGPT 4 | 0.392 | 0.350 | 0.440 | 0.394 | 0.558 | 0.459 | 0.827 | 0.834 |
| A3 – ChatGPT 4o | 0.626 | 0.620 | 0.526 | 0.393 | 0.741 | 0.475 | 0.865 | 0.898 |
| A4 - Gemini 1.5 | 0.333 | 0.181 | 0.388 | 0.276 | 0.610 | 0.558 | 0.435 | 0.463 |
| A5 - Claude 3.5 | 0.708 | 0.576 | 0.657 | 0.570 | 0.672 | 0.726 | 0.804 | 0.896 |
| A6 - OpenAI o1 | 0.422 | 0.302 | 0.557 | 0.369 | 0.543 | 0.304 | 0.656 | 0.584 |
| A7 – Gemini 2.0 | 0.201 | 0.131 | 0.319 | 0.199 | 0.289 | 0.143 | 0.716 | 0.619 |
| A8 – DeepSeek V3 | 0.378 | 0.233 | 0.610 | 0.439 | 0.619 | 0.448 | 0.664 | 0.667 |
| A9 - Gemini 1.5 pro | 0.267 | 0.169 | 0.698 | 0.576 | 0.631 | 0.640 | 0.590 | 0.537 |
| A10 - DeepSeek R1 | 0.579 | 0.522 | 0.417 | 0.293 | 0.453 | 0.368 | 0.727 | 0.703 |
| A11=Ensemble | 0.606 | 0.426 | 0.605 | 0.476 | 0.776 | 0.614 | 0.779 | 0.745 |
| A12=A11-A7 | 0.630 | 0.474 | 0.720 | 0.535 | 0.783 | 0.671 | 0.805 | 0.764 |
| A13=A12-A9 | 0.689 | 0.502 | 0.630 | 0.516 | 0.751 | 0.669 | 0.767 | 0.753 |
| A14=A13-A6 | 0.692 | 0.551 | 0.693 | 0.474 | 0.733 | 0.623 | 0.837 | 0.809 |
| A15= A14-A8 | 0.665 | 0.513 | 0.630 | 0.516 | 0.681 | 0.685 | 0.815 | 0.790 |
| A16= A15-A2-A4 | 0.688 | 0.588 | 0.675 | 0.563 | 0.647 | 0.675 | 0.812 | 0.790 |

Table 2
Analytic (Task Completion) Scoring Accuracy (QWK) for AI Raters against Each Human Rater

| Essay | SN1 | | SN2 | | ER1 | | ER2 | |
|---|---|---|---|---|---|---|---|---|
| Rater | R1 | R2 | R1 | R2 | R1 | R2 | R1 | R2 |
| R1/R2 | 0.715 | | 0.808 | | 0.776 | | 0.915 | |
| A1 – ChatGPT 3.5 | 0.523 | 0.577 | 0.673 | 0.533 | 0.543 | 0.514 | 0.624 | 0.674 |
| A2 – ChatGPT 4 | 0.305 | 0.422 | 0.600 | 0.451 | 0.512 | 0.525 | 0.832 | 0.825 |
| A3 – ChatGPT 4o | 0.593 | 0.645 | 0.685 | 0.551 | 0.607 | 0.716 | 0.893 | 0.892 |
| A4 - Gemini 1.5 | 0.171 | 0.223 | 0.427 | 0.319 | 0.635 | 0.654 | 0.569 | 0.587 |
| A5 - Claude 3.5 | 0.574 | 0.634 | 0.639 | 0.575 | 0.653 | 0.757 | 0.754 | 0.814 |
| A6 - OpenAI o1 | 0.258 | 0.307 | 0.568 | 0.427 | 0.385 | 0.475 | 0.569 | 0.599 |
| A7 – Gemini 2.0 | 0.341 | 0.226 | 0.444 | 0.407 | 0.287 | 0.328 | 0.631 | 0.717 |
| A8 – DeepSeek V3 | 0.380 | 0.251 | 0.525 | 0.372 | 0.580 | 0.580 | 0.651 | 0.650 |
| A9 - Gemini 1.5 pro | 0.344 | 0.121 | 0.492 | 0.456 | 0.696 | 0.729 | 0.621 | 0.644 |
| A10 - DeepSeek R1 | 0.324 | 0.477 | 0.459 | 0.387 | 0.576 | 0.620 | 0.722 | 0.786 |
| A11=Ensemble | 0.460 | 0.474 | 0.698 | 0.615 | 0.734 | 0.815 | 0.725 | 0.785 |
| A12=A11-A7 | 0.490 | 0.521 | 0.756 | 0.683 | 0.702 | 0.782 | 0.730 | 0.791 |
| A13=A12-A9 | 0.469 | 0.573 | 0.746 | 0.632 | 0.702 | 0.782 | 0.752 | 0.773 |
| A14=A13-A6 | 0.533 | 0.671 | 0.756 | 0.683 | 0.695 | 0.748 | 0.765 | 0.788 |
| A15= A14-A8 | 0.465 | 0.562 | 0.730 | 0.585 | 0.755 | 0.786 | 0.785 | 0.813 |
| A16= A15-A2-A4 | 0.481 | 0.699 | 0.690 | 0.566 | 0.676 | 0.767 | 0.735 | 0.819 |

Tables 2 to 4 summarize the analytic score accuracy in terms of QWK of each AI rater against each human rater on each trait: Task Completion, Delivery, and Language Use respectively. For analytic scores on Task Completion (Table 2), the two human rater had higher consistency compared with the holistic scoring, with QWK all above 0.7. For SN1, ChatGPT 4o yielded the highest QWK with both human raters on SN1 and ER2 as well as on SN2 with R1. Claude 3.5 yielded the highest QWK with R2 on both SN2 and ER1. For ER1, Gemini 1.5 Pro yielded the highest QWK with R1. Multiple ensemble models improved scoring accuracy over the standalone LLM for SN1, SN2, and ER1. Human-human QWK was consistently higher than AI-human QWK on each of the analytic scores.

For analytic scores on Delivery (Table 3), the two human rater yielded QWK above 0.7 on ER essays only. ChatGPT 4o yielded the highest QWK with both human raters on ER2. Claude 3.5 yielded the highest QWK with each of the two raters on both SN1 and ER1. For SN2, Gemini 1.5 Pro yielded the highest QWK with both human raters. Only two ensemble models improved scoring accuracy over the standalone LLMS for SN1 and ER1. Unlike analytic scores for Task Completion, AI-human QWK was consistently higher than human-human QWK on SN tasks and ER1 task except ER2.

For analytic scores on Language Use (Table 4), the two human rater yielded QWK above 0.7 on all tasks except SN2. ChatGPT 4.0 and 4o yielded the highest QWK with R1 and R2 respectively on ER2. Gemini 1.5 yielded the highest QWK with both raters on ER1. Claude 3.5 yielded the highest QWK with both raters on SN1. For SN2, Gemini 1.5 Pro yielded the highest QWK with both human raters. Only two ensemble models improved scoring accuracy over the standalone LLMS for ER1. AI-human QWK was higher than human-human QWK only on SN2.

Table 3

Analytic (Delivery) Scoring Accuracy (QWK) for AI Raters against Each Human Rater

| Essay | SN1 | | SN2 | | ER1 | | ER2 | |
|---|---|---|---|---|---|---|---|---|
| Rater | R1 | R2 | R1 | R2 | R1 | R2 | R1 | R2 |
| R1/R2 | 0.685 | | 0.588 | | 0.763 | | 0.950 | |
| A1 – ChatGPT 3.5 | 0.552 | 0.478 | 0.470 | 0.381 | 0.486 | 0.499 | 0.655 | 0.672 |
| A2 – ChatGPT 4 | 0.438 | 0.405 | 0.508 | 0.437 | 0.420 | 0.422 | 0.791 | 0.766 |
| A3 – ChatGPT 4o | 0.607 | 0.552 | 0.534 | 0.442 | 0.537 | 0.460 | 0.907 | 0.890 |
| A4 - Gemini 1.5 | 0.166 | 0.187 | 0.489 | 0.282 | 0.614 | 0.544 | 0.362 | 0.368 |
| A5 - Claude 3.5 | 0.698 | 0.599 | 0.660 | 0.614 | 0.652 | 0.781 | 0.746 | 0.803 |
| A6 - OpenAI o1 | 0.265 | 0.284 | 0.480 | 0.342 | 0.521 | 0.344 | 0.593 | 0.599 |
| A7 – Gemini 2.0 | 0.174 | 0.117 | 0.219 | 0.173 | 0.237 | 0.197 | 0.468 | 0.500 |
| A8 – DeepSeek V3 | 0.346 | 0.286 | 0.424 | 0.436 | 0.619 | 0.468 | 0.544 | 0.597 |
| A9 - Gemini 1.5 pro | 0.361 | 0.221 | 0.681 | 0.631 | 0.497 | 0.463 | 0.731 | 0.718 |
| A10 - DeepSeek R1 | 0.532 | 0.522 | 0.194 | 0.244 | 0.526 | 0.471 | 0.577 | 0.632 |
| A11=Ensemble | 0.541 | 0.417 | 0.545 | 0.444 | 0.667 | 0.596 | 0.670 | 0.723 |
| A12=A11-A7 | 0.548 | 0.438 | 0.622 | 0.499 | 0.720 | 0.597 | 0.692 | 0.745 |
| A13=A12-A9 | 0.593 | 0.486 | 0.571 | 0.526 | 0.744 | 0.588 | 0.700 | 0.756 |
| A14=A13-A6 | 0.609 | 0.561 | 0.570 | 0.526 | 0.695 | 0.647 | 0.739 | 0.802 |
| A15= A14-A8 | 0.594 | 0.527 | 0.574 | 0.508 | 0.721 | 0.712 | 0.768 | 0.784 |
| A16= A15-A2-A4 | 0.689 | 0.650 | 0.541 | 0.449 | 0.731 | 0.684 | 0.789 | 0.828 |

Table 4

Analytic (Language Use) Scoring Accuracy (QWK) for AI Raters against Each Human Rater

| Essay | SN1 | | SN2 | | ER1 | | ER2 | |
|---|---|---|---|---|---|---|---|---|
| Rater | R1 | R2 | R1 | R2 | R1 | R2 | R1 | R2 |
| R1/R2 | 0.832 | | 0.488 | | 0.783 | | 0.939 | |
| A1 – ChatGPT 3.5 | 0.462 | 0.453 | 0.503 | 0.459 | 0.420 | 0.500 | 0.560 | 0.603 |
| A2 – ChatGPT 4 | 0.409 | 0.433 | 0.424 | 0.337 | 0.446 | 0.422 | 0.852 | 0.829 |
| A3 – ChatGPT 4o | 0.366 | 0.348 | 0.275 | 0.213 | 0.376 | 0.375 | 0.705 | 0.846 |
| A4 - Gemini 1.5 | 0.235 | 0.246 | 0.496 | 0.293 | 0.695 | 0.677 | 0.380 | 0.416 |
| A5 - Claude 3.5 | 0.667 | 0.686 | 0.585 | 0.596 | 0.583 | 0.594 | 0.715 | 0.744 |
| A6 - OpenAI o1 | 0.236 | 0.283 | 0.451 | 0.397 | 0.341 | 0.252 | 0.553 | 0.581 |
| A7 – Gemini 2.0 | 0.140 | 0.124 | 0.017 | 0.104 | 0.221 | 0.139 | 0.495 | 0.528 |
| A8 – DeepSeek V3 | 0.227 | 0.074 | 0.265 | 0.246 | 0.368 | 0.417 | 0.553 | 0.581 |
| A9 - Gemini 1.5 pro | 0.447 | 0.304 | 0.626 | 0.677 | 0.599 | 0.622 | 0.495 | 0.528 |
| A10 - DeepSeek R1 | 0.432 | 0.382 | 0.216 | 0.246 | 0.455 | 0.459 | 0.665 | 0.743 |
| A11=Ensemble | 0.460 | 0.382 | 0.429 | 0.417 | 0.644 | 0.558 | 0.618 | 0.685 |
| A12=A11-A7 | 0.460 | 0.382 | 0.560 | 0.455 | 0.672 | 0.709 | 0.652 | 0.716 |
| A13=A12-A9 | 0.432 | 0.364 | 0.509 | 0.397 | 0.690 | 0.602 | 0.688 | 0.761 |
| A14=A13-A6 | 0.462 | 0.389 | 0.536 | 0.421 | 0.594 | 0.641 | 0.701 | 0.773 |
| A15= A14-A8 | 0.492 | 0.443 | 0.551 | 0.413 | 0.729 | 0.748 | 0.685 | 0.741 |
| A16= A15-A2-A4 | 0.618 | 0.598 | 0.497 | 0.416 | 0.670 | 0.659 | 0.730 | 0.783 |

For scoring accuracy, no AI rater turned out to be always the best performer across all tasks. In general, ChatGPT 4o, Claude 3.5, and Gemini 1.5 Pro yielded the highest performing AI raters depending on the scoring tasks, holistic or analytic, as well as the specific trait scores. Ensemble models were able to occasionally improve the scoring accuracy for a specific essay compared with the standalone LLMs.

**Intra-Rater Consistency**

This study further examined the intra-rater consistency using the Cronbach Alpha as reported in Table 5. The number of items used for computing intra-rater consistency is 4 items for the holistic scores and analytic scores for each trait respectively, 12 items for analytic scores across all traits, and 6 items for analytic scores for each type of essays: SN and ER. The intra-rater consistency was higher when the number of items included for the analysis was higher. Again, the top performing AI raters, either ChatGPT 4o or Claude 3.5, still yielded the highest intra-rater consistency across different tasks, yielding slightly lower values on Language Use trait. The intra-rater consistency for human rater R2 was consistently higher than that for human rater R1. The intra-rater consistency for the best performing AI raters was about the same or higher than R2 except for analytic scores of Language Use and ER essay.

Table 5
Intra-Rater Consistency

|  | Holistic | Trait-Task Completion | Trait-Delivery | Trait-Language Use | Analytic | SN | ER |
|---|---|---|---|---|---|---|---|
| # of Items | 4 | 4 | 4 | 4 | 12 | 6 | 6 |
| R1 | 0.71 | 0.73 | 0.67 | 0.58 | 0.90 | 0.89 | 0.91 |
| R2 | 0.81 | 0.81 | 0.78 | 0.77 | 0.94 | 0.95 | 0.93 |
| A1 – ChatGPT 3.5 | 0.71 | 0.66 | 0.73 | 0.72 | 0.92 | 0.91 | 0.89 |
| A2 – ChatGPT 4 | 0.66 | 0.64 | 0.64 | 0.58 | 0.89 | 0.92 | 0.79 |
| A3 – ChatGPT 4o | 0.81 | 0.80 | 0.79 | 0.58 | 0.91 | 0.90 | 0.90 |
| A4 - Gemini 1.5 | 0.46 | 0.60 | 0.63 | 0.58 | 0.88 | 0.90 | 0.86 |
| A5 - Claude 3.5 | 0.80 | 0.82 | 0.78 | 0.74 | 0.94 | 0.95 | 0.89 |
| A6 - OpenAI o1 | 0.67 | 0.62 | 0.50 | 0.56 | 0.87 | 0.85 | 0.79 |
| A7 – Gemini 2.0 | 0.35 | 0.26 | 0.28 | 0.30 | 0.72 | 0.67 | 0.80 |
| A8 – DeepSeek V3 | 0.62 | 0.69 | 0.51 | 0.58 | 0.87 | 0.92 | 0.82 |
| A9 - Gemini 1.5 pro | 0.61 | 0.61 | 0.60 | 0.41 | 0.85 | 0.86 | 0.81 |
| A10 - DeepSeek R1 | 0.68 | 0.72 | 0.73 | 0.67 | 0.91 | 0.89 | 0.89 |

**AI Rater Effects**

The mean scores of the holistic and analytic scores assigned by each rater are summarized in Tables 6. The standard deviation of the scores assigned are presented in Table B.1 in Appendix B. The red colored numbers indicate the AI rater who assigned scores with the highest mean score among all AI raters or a human rater with higher scores compared with the other human rater while the blue-colored numbers indicate the lowest mean scores assigned among AI raters. To some degree, this pattern reflects the patterns observed in the rater effects of severity vs leniency. Further, Table 7 presents the standard deviation of the scores assigned by each rater, which may help better understand the rater centrality effect. In general, human rater R2 assigned higher mean scores across all rating tasks compared with R1. Geimini 2.0 assigned the lowest means cores on majority of the rating tasks while ChatGPT 3.5 and ChatGPT 4 produced the highest mean scores on majority of the rating tasks. The pattern in standard deviation of the scores is not clear.

The Wright Maps from the FACET models are presented in Figures 1 and 2 for the holistic and the analytic scores across three traits. Figures 3 to 5 present the Wright Map for the analytic scores for each trait. These maps demonstrate the relative locations of examinees, essays, and raters' parameters on the latent trait scale ranging from -3 to +3. For holistic scoring as shown in Figure 1, SN2 was the most difficult essay, while ER2 was the easiest essay. Gemini 2.0 was the most stringent rater, which is consistent with the lowest mean scores assigned while R2 and ChatGPT 3.5 were more lenient raters as they assigned the highest or higher mean scores as presented in Table 6.

Table 7 presents the rater parameter estimates and fit measures of each rater to the FACET model. Following equation 1, when the rater parameter is 0, the rater has no effect on scoring. A positive rater parameter means the likelihood of getting a score will be decreased due to a severe rater effect while a negative rater parameter means the likelihood of getting a score will be increased due to a lenient rater effect. Human rater R2 was more lenient than R1, with a logit value difference of about 0.6 which can be considered as non-negligible difference if we follow the cut value of 0.3 logit (Miller, Rotou, & Twing, 2004) for parameter drift evaluation.

Table 6

Mean Holistic and Analytic Scores Assigned by Each Rater

| Scores | R1 | R2 | A1 | A2 | A3 | A4 | A5 | A6 | A7 | A8 | A9 | A10 | Average |
|---|---|---|---|---|---|---|---|---|---|---|---|---|---|
| SN1-Holistic | 3.70 | 4.10 | 3.53 | 3.03 | 3.73 | 2.83 | 3.80 | 3.17 | 2.60 | 3.67 | 3.33 | 3.50 | 3.42 |
| ER1-Holistic | 3.30 | 3.83 | 4.30 | 3.43 | 2.93 | 3.43 | 3.50 | 2.90 | 2.37 | 3.10 | 3.80 | 3.57 | 3.37 |
| SN2-Holistic | 3.23 | 3.70 | 3.50 | 2.97 | 2.70 | 2.90 | 3.10 | 2.87 | 2.00 | 3.47 | 3.47 | 3.37 | 3.11 |
| ER2-Holistic | 3.73 | 4.07 | 3.13 | 3.93 | 3.77 | 3.60 | 3.87 | 3.10 | 3.10 | 3.47 | 3.20 | 3.50 | 3.54 |
| Task_SN1 | 4.20 | 4.03 | 3.70 | 3.30 | 4.23 | 2.83 | 3.80 | 3.23 | 3.57 | 4.30 | 3.70 | 3.30 | 3.68 |
| Task_SN2 | 3.60 | 3.73 | 3.53 | 3.33 | 3.47 | 2.90 | 3.03 | 3.03 | 3.73 | 4.13 | 3.77 | 3.47 | 3.48 |
| Task_ER1 | 3.93 | 3.70 | 4.17 | 3.87 | 3.30 | 3.40 | 3.57 | 3.17 | 2.57 | 3.30 | 4.00 | 3.63 | 3.55 |
| Task_ER2 | 4.03 | 3.87 | 3.27 | 3.93 | 3.87 | 3.57 | 3.60 | 3.03 | 3.27 | 3.43 | 3.27 | 3.53 | 3.56 |
| Delivery_SN1 | 3.77 | 4.03 | 3.43 | 3.03 | 3.50 | 2.67 | 3.80 | 2.90 | 2.70 | 3.60 | 3.33 | 3.43 | 3.35 |
| Delivery_SN2 | 3.23 | 3.63 | 3.43 | 3.10 | 2.73 | 2.90 | 3.07 | 2.87 | 2.40 | 3.47 | 3.47 | 2.93 | 3.10 |
| Delivery_ER1 | 3.53 | 3.73 | 4.30 | 3.87 | 2.97 | 3.57 | 3.97 | 2.97 | 2.43 | 3.40 | 4.27 | 3.43 | 3.54 |
| Delivery_ER2 | 3.97 | 3.87 | 3.20 | 4.03 | 3.87 | 3.63 | 3.67 | 3.03 | 3.37 | 3.23 | 3.57 | 3.43 | 3.57 |
| LUSE_SN1 | 3.83 | 3.97 | 3.47 | 3.03 | 2.97 | 2.77 | 3.80 | 3.10 | 2.40 | 3.10 | 3.27 | 3.27 | 3.25 |
| LUSE_SN2 | 3.30 | 3.53 | 3.43 | 2.87 | 2.40 | 2.90 | 3.03 | 2.80 | 2.20 | 2.50 | 3.53 | 2.83 | 2.94 |
| LUSE_ER1 | 3.43 | 3.60 | 4.33 | 3.37 | 2.67 | 3.40 | 3.23 | 2.63 | 2.37 | 2.80 | 3.97 | 3.13 | 3.24 |
| LUSE_ER2 | 4.10 | 3.90 | 3.20 | 3.90 | 3.77 | 3.53 | 3.63 | 3.03 | 3.17 | 3.03 | 3.17 | 3.40 | 3.49 |

Note: A1 – ChatGPT 3.5, A2 – ChatGPT 4, A3 – ChatGPT 4o, A4 - Gemini 1.5, A5 - Claude 3.5 Sonnet, A6 - OpenAI o1, A7 – Gemini 2.0, A8 – DeepSeek V3, A9 - Gemini 1.5 Pro, A10 - DeepSeek R1.

Consistent with patterns of the mean scores in Table 6 and what presented in Figure 1, Gemini 2.0 was an extreme severe rater with a logit value of 1.25. ChatGPT 4 was the most neutral rater with a rater parameter estimate close to 0. If we follow the cut value of 0.3, all AI raters did not have severe rater effects except ChatGPT 3.5 (-0.37) at the lenient end, while OpenAI o1 (0.5) at the harsh end. The two human raters had the smallest infit and outfit mean squares (both smaller than the cut value of 0.7), indicating potential central tendency effect. DeepSeek R1, ChatGPT 3.5, and Gemini 1.5 Pro had infit and outfit mean squares larger than 1.3, indicating potential misfit.

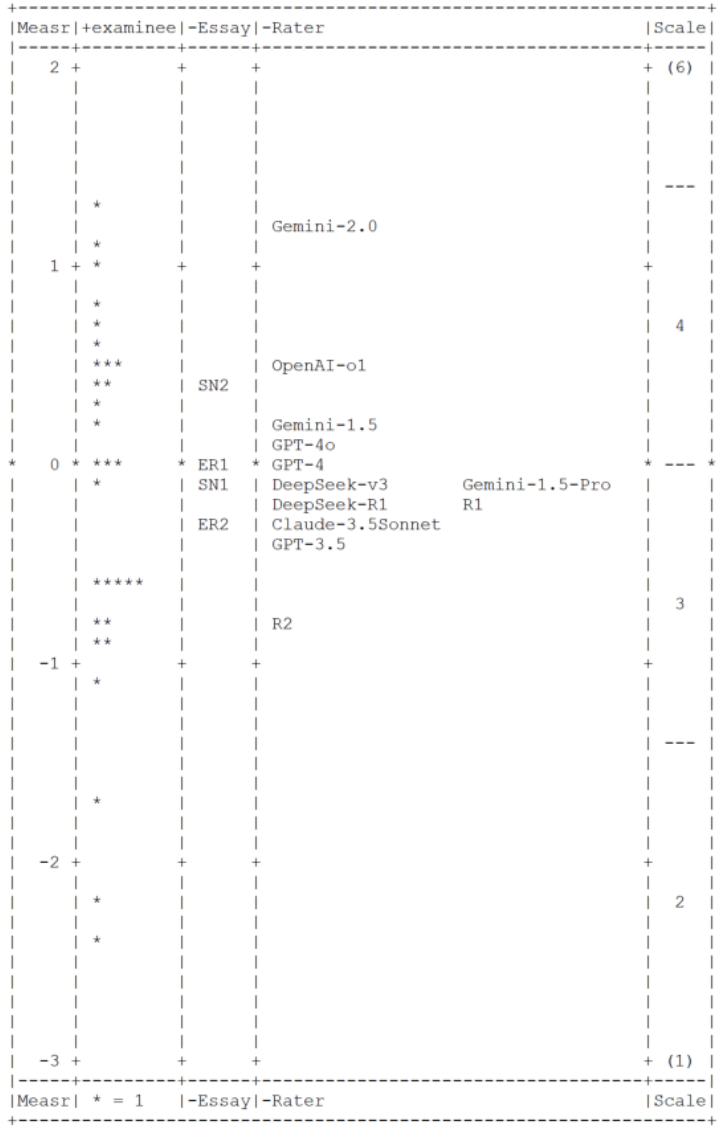

Figure 1. The Wright Map for the Holistic Scores

For analytic scoring across all three traits as shown in Figure 2, SN2 was still the most difficult essay, while ER2 was the easiest essay. SN1 was the second difficult essay. Gemini 2.0 was the most stringent rater, which is consistent with the lowest mean scores assigned while two human raters were the most lenient raters. Among the three traits, Language Use was the most difficult, while Task Completion was the easiest.

Table 8 presents the rater parameter estimates for the analytic scores. Human rater R2 was slightly more lenient than R1, but with a smaller difference of about 0.08. Consistent with what presented in Figure 2, Gemini 2.0 was a severe rater with a logit value of 0.76. DeepSeek V3 was the most neutral rater with a rater parameter estimate close to 0. All AI raters did not have severe rater effects except OpenAI o1 (0.5) at the harsh end. OpenAI o1 had the least infit and outfit mean squares at the borderline (0.69). No infit and outfit mean squares were larger than 1.3 for any rater.

Table 7
Rater Parameter Estimates for Holistic Scoring

| Rater | Rater Par | SE | Infit MS | Outfit MS |
| --- | --- | --- | --- | --- |
| 2 R2 | -0.81 | 0.11 | 0.60 | 0.60 |
| 3 ChatGPT 3.5 | -0.37 | 0.11 | 1.36 | 1.36 |
| 7 Claude 3.5 Sonnet | -0.30 | 0.11 | 1.01 | 1.01 |
| 1 R1 | -0.20 | 0.11 | 0.58 | 0.59 |
| 12 DeepSeek R1 | -0.18 | 0.11 | 1.37 | 1.41 |
| 11 Gemini 1.5 Pro | -0.14 | 0.11 | 1.41 | 1.41 |
| 10 DeepSeek V3 | -0.10 | 0.11 | 1.07 | 1.08 |
| 4 ChatGPT 4 | 0.02 | 0.11 | 0.91 | 0.92 |
| 5 ChatGPT 4o | 0.10 | 0.11 | 0.85 | 0.84 |
| 6 Gemini 1.5 | 0.23 | 0.11 | 0.85 | 0.88 |
| 8 OpenAI o1 | 0.50 | 0.11 | 0.70 | 0.69 |
| 9 Gemini 2.0 | 1.25 | 0.11 | 1.22 | 1.29 |

Table 8
Rater Parameter Estimates for Analytic Scoring

| Rater | Rater Par | SE | Infit MS | Outfit MS |
| --- | --- | --- | --- | --- |
| 2 R2 | -0.55 | 0.06 | 0.73 | 0.73 |
| 1 R1 | -0.47 | 0.06 | 0.83 | 0.84 |
| 3 ChatGPT 3.5 | -0.31 | 0.06 | 1.21 | 1.20 |
| 11 Gemini 1.5 Pro | -0.29 | 0.06 | 1.19 | 1.19 |
| 7 Claude 3.5 Sonnet | -0.17 | 0.06 | 0.97 | 0.97 |
| 4 ChatGPT 4 | -0.10 | 0.06 | 0.88 | 0.89 |
| 10 DeepSeek-v3 | 0.05 | 0.06 | 1.18 | 1.18 |
| 12 DeepSeek R1 | 0.10 | 0.06 | 1.15 | 1.18 |
| 5 ChatGPT 4o | 0.11 | 0.06 | 1.06 | 1.05 |
| 6 Gemini 1.5 | 0.30 | 0.06 | 0.84 | 0.87 |
| 8 OpenAI o1 | 0.57 | 0.06 | 0.69 | 0.69 |
| 9 Gemini 2.0 | 0.76 | 0.06 | 1.21 | 1.26 |

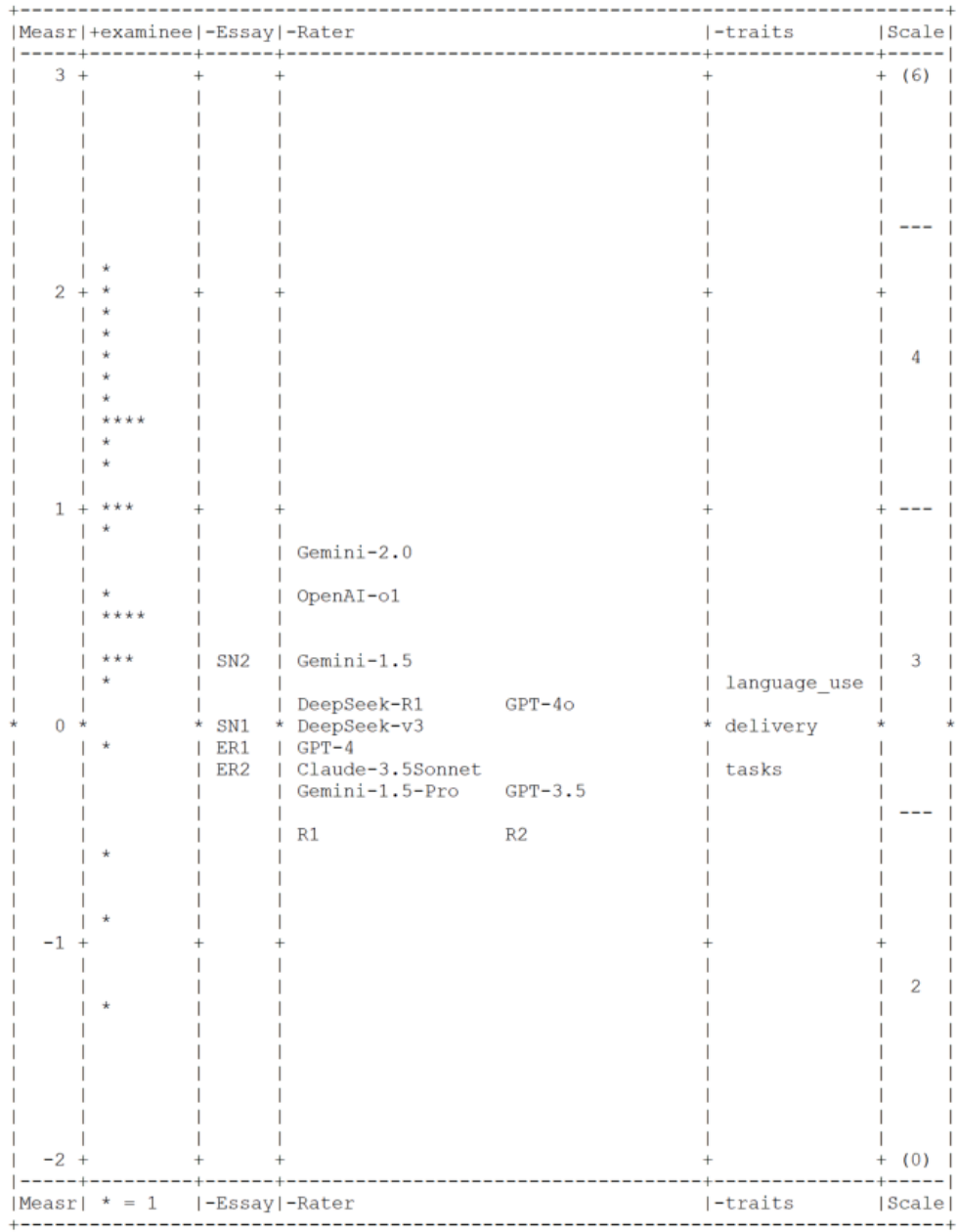

Figure 2. The Wright Map for the Analytic Scores

Table 9 presents the rater parameter estimates for the analytic scores for Task Completion. Two human raters were most lenient with R1 slightly more lenient than R2. OpenAI o1 was the most severe rater with a logit value of 0.61. ChatGPT 4 and Claude 3.5 were more neutral rater with a rater parameter estimate smaller than an absolute value of 0.1. All AI raters did not have severe rater effects except OpenAI o1 (0.5) at the harsh end, followed by Gemini 1.5 (0.53) and 2.0 (0.38). No infit and outfit mean squares were smaller than 0.7 or larger than 1.3 for any rater.

Figures 3, 4, and 5 present the Wright Maps for the analytic scores assigned to each of the three traits. For the Task Completion trait (Figure 3), SN2 was still the most difficult essay while SN1 was the easiest essay. ER1 and ER2 had about the same difficulty. OpenAI o1 was the most stringent rater while two human raters were the most lenient raters. R1 was more lenient than R2 on Task Completion trait scoring.

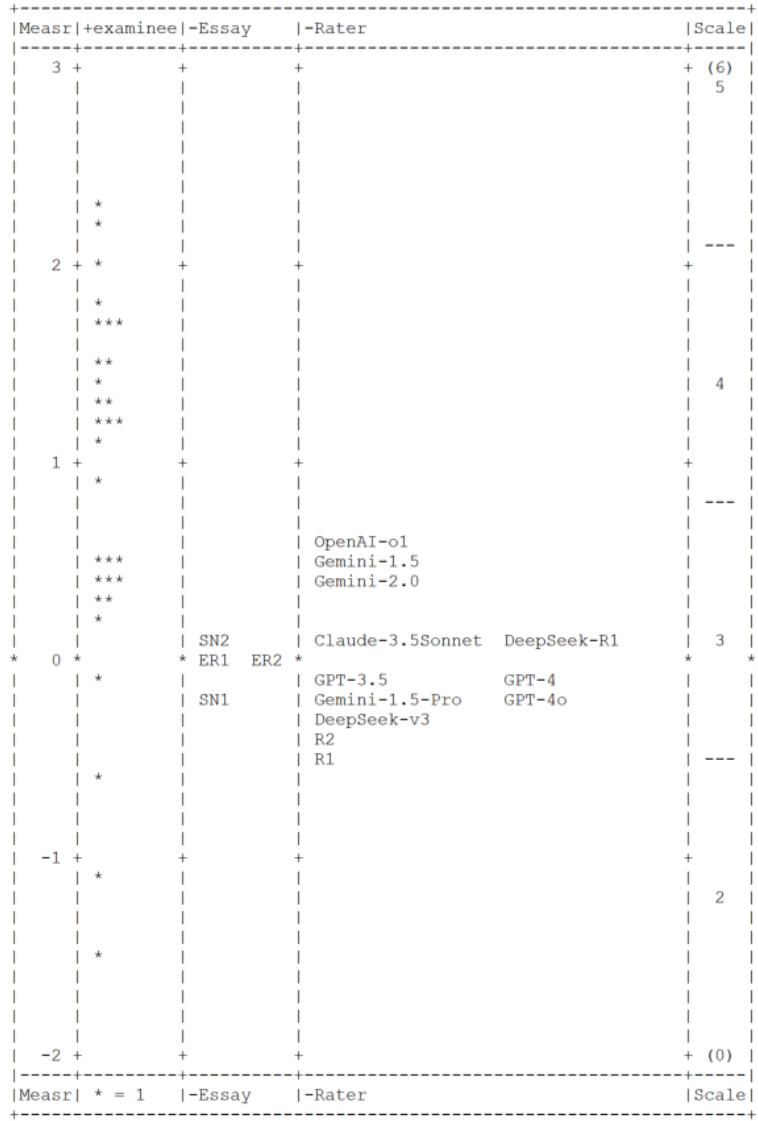

Figure 3. The Wright Map for the Analytic Scores-Task Completion

For the Delivery trait (Figure 4), SN2 was still the most difficult essay while ER2 was the easiest essay. Gemini 2.0 was the most stringent rater while R2 was the most lenient raters. R1 was the second most lenient rater like Claude 3.5 and Gemini 1.5 Pro on Delivery trait scoring.

Table 9
Rater Parameter Estimates for Analytic Scoring-Task Completion

| Rater | Rater Par | SE | Infit MS | Outfit MS |
| --- | --- | --- | --- | --- |
| 1 R1 | -0.51 | 0.11 | 0.97 | 0.96 |
| 2 R2 | -0.36 | 0.11 | 0.74 | 0.73 |
| 10 DeepSeek V3 | -0.30 | 0.11 | 1.24 | 1.23 |
| 5 ChatGPT 4o | -0.20 | 0.11 | 0.97 | 0.96 |
| 11 Gemini 1.5 Pro | -0.16 | 0.11 | 1.24 | 1.23 |
| 3 ChatGPT 3.5 | -0.13 | 0.11 | 1.16 | 1.16 |
| 4 ChatGPT 4 | -0.06 | 0.11 | 0.90 | 0.88 |
| 7 Claude 3.5 Sonnet | 0.09 | 0.11 | 0.96 | 0.96 |
| 12 DeepSeek R1 | 0.11 | 0.11 | 1.15 | 1.15 |
| 9 Gemini 2.0 | 0.38 | 0.11 | 1.04 | 1.05 |
| 6 Gemini 1.5 | 0.53 | 0.11 | 0.85 | 0.88 |
| 8 OpenAI o1 | 0.61 | 0.11 | 0.78 | 0.78 |

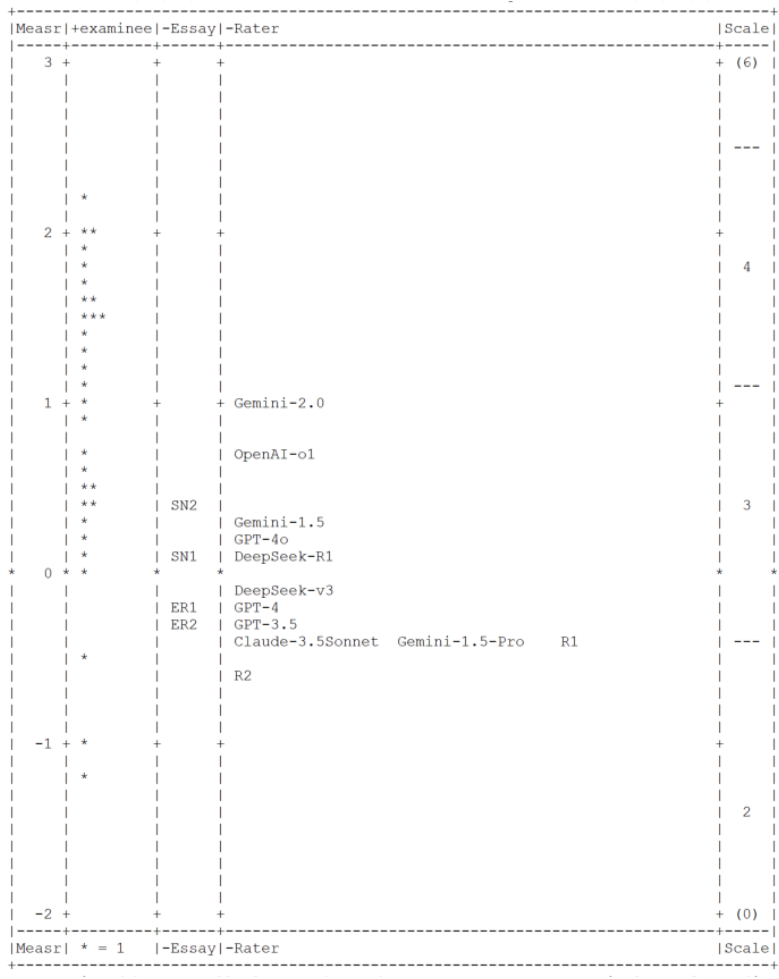

Figure 4. The Wright Map for the Analytic Scores-Delivery

Table 10
Rater Parameter Estimates for Analytic Scoring-Delivery

| Rater | Rater Par | SE | Infit MS | Outfit MS |
|---|---|---|---|---|
| 2 R2 | -0.63 | 0.11 | 0.78 | 0.80 |
| 11 Gemini 1.5 Pro | -0.40 | 0.11 | 1.13 | 1.12 |
| 1 R1 | -0.35 | 0.11 | 0.80 | 0.81 |
| 7 Claude 3.5 Sonnet | -0.35 | 0.11 | 1.01 | 1.01 |
| 3 ChatGPT 3.5 | -0.30 | 0.11 | 1.23 | 1.22 |
| 4 ChatGPT 4 | -0.18 | 0.11 | 0.93 | 0.95 |
| 10 DeepSeek V3 | -0.06 | 0.11 | 1.09 | 1.09 |
| 12 DeepSeek R1 | 0.11 | 0.11 | 1.29 | 1.32 |
| 5 ChatGPT 4o | 0.18 | 0.11 | 0.99 | 0.99 |
| 6 Gemini 1.5 | 0.29 | 0.11 | 0.87 | 0.90 |
| 8 OpenAI o1 | 0.68 | 0.11 | 0.70 | 0.71 |
| 9 Gemini 2.0 | 1.03 | 0.12 | 1.07 | 1.13 |

Table 10 presents the rater parameter estimates for the analytic scores for Delivery. Human rater R2 was the most lenient. Gemini 2.0 was the most severe rater with a rater parameter of 1.03. DeepSeek V3 was the most neutral rater with a rater parameter estimate of -0.06. Using an absolute cut value of 0.3, OpenAI o1 was a harsh rater, while Claude 3.5 was slightly lenient. All other AI raters did not have severe rater effects. No infit and outfit mean squares were smaller than 0.7 or larger than 1.3 for any rater.

Table 11
Rater Parameter Estimates for Analytic Scoring-Language Use

| Rater | Rater Par | SE | Infit MS | Outfit MS |
|---|---|---|---|---|
| 2 R2 | -0.74 | 0.11 | 0.72 | 0.74 |
| 1 R1 | -0.63 | 0.11 | 0.78 | 0.79 |
| 3 ChatGPT 3.5 | -0.55 | 0.11 | 1.33 | 1.33 |
| 11 Gemini 1.5 Pro | -0.37 | 0.11 | 1.23 | 1.23 |
| 7 Claude 3.5 Sonnet | -0.29 | 0.11 | 0.96 | 0.96 |
| 4 ChatGPT 4 | -0.10 | 0.11 | 0.90 | 0.92 |
| 12 DeepSeek R1 | 0.09 | 0.11 | 1.09 | 1.14 |
| 6 Gemini 1.5 | 0.10 | 0.11 | 0.88 | 0.92 |
| 5 ChatGPT 4o | 0.40 | 0.11 | 1.10 | 1.09 |
| 8 OpenAI o1 | 0.49 | 0.11 | 0.64 | 0.63 |
| 10 DeepSeek V3 | 0.54 | 0.11 | 0.97 | 0.96 |
| 9 Gemini 2.0 | 1.06 | 0.12 | 1.39 | 1.42 |

For the Language Use trait (Figure 5), SN2 was still the most difficult essay while ER2 was the easiest essay. Gemini 2.0 was the most stringent rater while rater 2 was the most lenient raters. R1 was the second most lenient rater on Language Use trait scoring.

Table 11 presents the rater parameter estimates for the analytic scores for Language Use. Human rater R2 was the most lenient, R1 the second most lenient. Gemini 2.0 was the most severe rater with a rater parameter of 1.06. DeepSeek R1, ChatGPT 4 and Gemini 1.5 were more neutral with an absolute rater parameter estimate around 0.1. Using the absolute cut value of 0.3, DeepSeek V3, OpenAI o1, and ChatGPT 4o were slightly harsh raters while Gemini 1.5 Pro and ChatGPT 3.5 were slightly lenient. OpenAI o1 had infit and outfit mean squares smaller than 0.7, indicating slight central tendency, while ChatGPT 3.5 and Gemini 2.0 yielded infit and outfit mean squares larger than 1.3, indicating slight misfit.

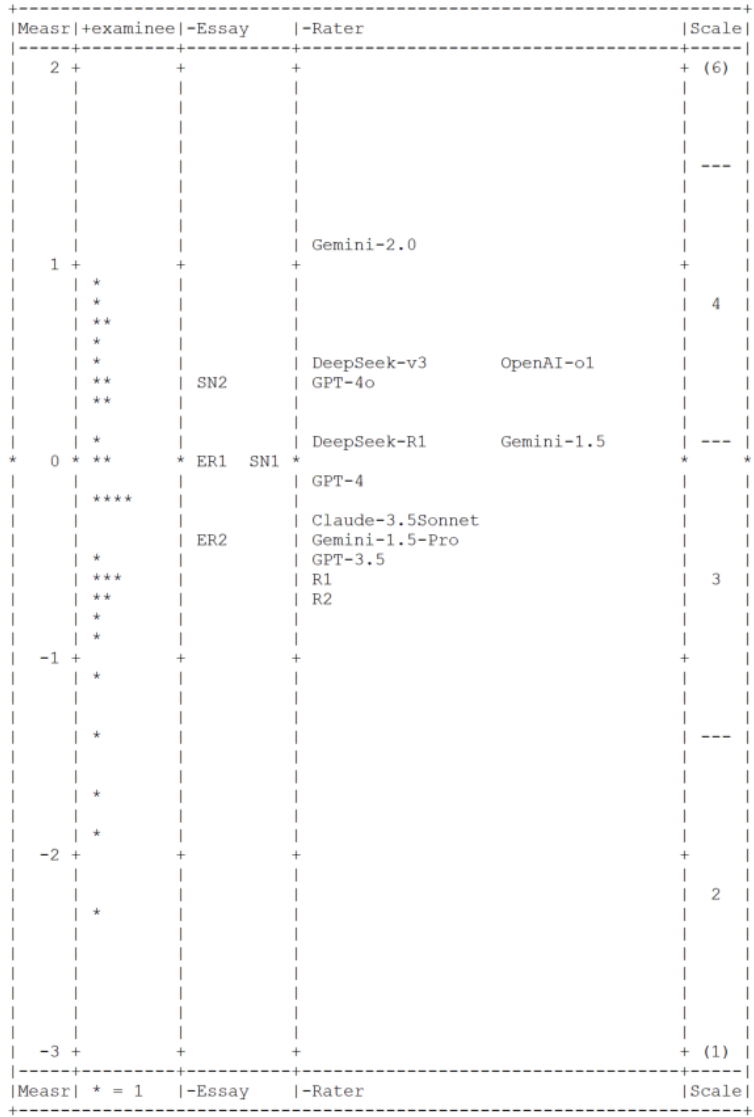

Figure 5. The Wright Map for the Analytic Scores-Language Use

## Summary and Discussion

This study aims to examine the rater effects of LLM-based AI raters in automated scoring. To illustrate the methods for exploring AI rater effects, this study investigated the use of LLMs for automated holistic and analytic scoring using a fully crossed rating design. Ten LLMs were compared including ChatGPT 3.5, ChatGPT 4, ChatGPT 4o, OpenAI o1, Claude 3.5 Sonnet, Gemini 1.5, Gemini 1.5 Pro, Gemini 2.0, as well as DeepSeek V3, and DeepSeek R1 against two human expert raters in scoring two types of writing tasks for the AP Chinese exams. The holistic and analytic score accuracy of LLMs compared with human raters were evaluated in terms of QWK. Intra-rater consistency across essays was compared in terms of Cronbach Alpha. Rater effects of LLMs were evaluated and compared with human raters using the Many FACET Rasch model.

For Research Question 1 related to the AI raters' accuracy, for holistic scoring, ChatGPT 4o, Claude 3.5, and Gemini 1.5 Pro yielded the highest scoring accuracy against either human rater. ChatGPT 4o performed the best on ER items, while Gemini 1.5 Pro excelled on SN items while Claude 3.5 showed strong performance on both SN and ER items. These three LLMs turned out to be the best performing models in analytic scoring as well. For Task Completion, ChatGPT 4o performed the best for SN and ER2 items while Claude 3.5 and Gemini 1.5 Pro performed the best on SN2 and ER1 essays. For Delivery, Claude 3.5 performed the best on SN1 and ER1 while ChatGPT4o and Gemini 1.5 Pro performed the best on ER2 and SN2 respectively. For the Language Use trait, ChatGPT 4, 4o, and Gemini 1.5 were the top performers on ER essays while Claude 3.5 and Gemini 1.5 Pro were the best performers for SN essays. In general, QWKs for the best performing AI raters were higher than those between two human raters for holistic scoring. However, human-human QWK were higher on analytic scoring than AI QWKs. QWKs for analytic scores were higher than those for holistic scoring.

For Research Question 2 related to the AI raters' intra-rater consistency, ChatGPT 4o yielded the highest intra-rater consistency for the holistic scores, the analytic scores on the Delivery trait, and the ER items. On the other hand, Claude 3.5 yielded the highest consistency for the analytic scores, analytic scores on the Task Completion and Language Use traits as well as the SN items. Human rater R2 yielded higher scoring consistency compared with R1. The best performing AI raters' intra-rater consistency were about the same as those for the best performing human rater except on ER essays and analytic scores on Language Use with slightly lower consistency.

For Research Questions 3 on the rater effects, human raters in general were more lenient. Gemini 2.0 turned out to be a harsh rater on most scoring tasks. ChatGPT 3.5 turned out to be a lenient rater in 3 out of 5 scoring analyses with DeepSeek V3 and Gemini 1.5 Pro as the two other lenient raters. Whether an AI rater was more neutral raters depends on the task. There was no evident central tendency effects for the AI raters. However, the two human raters displayed some minor central tendency effect in holistic scoring. As Wolfe (2004) pointed out, the rater effects estimated in the FACET Rasch model was scaled to a mean of 0. If the number of raters included in the comparison differs, the estimated rater effect may be different.

The results in general support the use of ChatGPT 4o, Gemini 1.5 Pro, and Claude 3.5 Sonnet with high scoring accuracy, better intra-rater consistency, and less rater effects. However, there were no AI raters which consistently performed the best. Some further explorations will be conducted to better understand the AI rater scoring logic. The findings of this study provide insights into the strengths and limitations of AI raters compared to human raters using the data in the study.

The limitations of the study lies in the sample size, the content domain, the type and the number of essays. There were only 30 students who participated in this experiment. The FACET software manual states a sample size of 30 as a minimum sample size. However, as the FACET Rasch model is an item response theory based latent trait model, a larger sample size would be expected even though the standard error of the rater parameter estimates was relatively small in the current study. Further, the content of the tasks was related to the AP Chinese exam. Whether the findings from this study is generalizable to other content domains such as English writing, reading, science, and math constructed-response items awaits further exploration. In addition, only two items types: SN and ER were included in the experiments, with a total of only 4 items. Other task types and traits for analytic scoring should be examined in future studies. The last limitation is related to the prompting methods for automated scoring using LLMs. The current study used a few-shot methods. Other methods should be explored in future studies including Chain-of-Thought and other fine-tuning methods.

In summary, this study assessed the scoring accuracy, intra-rater consistency of AI and human holistic and analytic ratings, and estimated the AI rater effects. It demonstrated the methods that can be used to examine AI rater performance, in particular, the use of the Many FACET Rasch model to evaluate rater severity/leniency and central tendency effects. This study provided empirical evidence to inform the adoption of LLM-based AI raters for automated holistic and analytic scoring. It is highlighted that when LLMs are used for automated scoring, the methods demonstrated in this study could be the starting point for such evaluation.

Appendix A

**AI Training Prompting for Grading SN1 Student Essays**

Based on this prompt, grading rubric, and student writing samples, grade student's writing sample.

Prompt: In this task, you will be asked to write in Chinese for a specific purpose and to a specific person. You should write in as complete and culturally appropriate a manner as possible, taking into account the purpose and the person described. The four pictures present a story. Imagine you are writing the story to a friend. Narrate a complete story as suggested by the pictures. Give your story a beginning, a middle, and an end.

**The first picture:** A teenage boy is lying on his bed. His room is messy, and his dad is asking him to clean it up. There are many things on the bed and on the floor, including a guitar, basketball, football, books, and more. He is listening to music.

**The second picture:** The boy puts everything under his bed, covers it with a bedsheet, and pretends he has cleaned the room.

**The third picture:** His dad removes the bedsheet, sees the mess, and is very surprised and angry.

**The fourth picture:** His dad goes downstairs and tells his son what he saw. The boy is sitting on the floor, playing video games, eating French fries, and drinking cola.

The task for you is to rate the student responses. Assess each response holistically based on three key criteria: task completion, delivery, and language use, with task completion being the most critical factor. Start by determining whether the student fully addresses all aspects of the four pictures. Then, evaluate delivery and language use before assigning an overall score. Additionally, please provide three analytic scores—one for each criterion—following the rubric's guidelines. When rating, refer to the sample responses for each score level as benchmarks. In total, provide four scores: one overall score and three analytic scores. Tips: Student responses will be graded based on task completion, delivery, and language use. However, it is important to apply a holistic grading approach. Among these criteria, task completion is the most important. If a student addresses only three pictures, they will receive 3 points, regardless of the quality of their language use and delivery. A student will receive 4 points if they discuss all four pictures. In this context, use grading rubrics 5 and 6 as references for evaluating language use and delivery. If a student addresses three pictures but demonstrates weak language use and delivery, they may receive 2 points. A student who addresses only one or two pictures but includes one or two sentences may also receive 2 points. If a student writes only isolated words related to the prompts (pictures), they may receive 1 point.

Grading Rubric:

Score of 0: UNACCEPTABLE: 1) Contains nothing that earns credit 2) Completely irrelevant to the stimulus 3) Not in Chinese characters, NR (No Response), BLANK (no response)

Samples:

Score 0: 你好

Score 0: I want to eat icecream.

Score of 1: Very Weak: Demonstrates lack of competence in interpersonal writing. - TASK COMPLETION: 1) E-mail addresses stimulus only minimally 2) Lacks organization and coherence; very disjointed sentences or isolated words -DELIVERY: 1) Constant use of register inappropriate to situation -LANGUAGE USE: 1) Insufficient, inappropriate vocabulary, with frequent errors that significantly obscure meaning; constant interference from another language 2) Little or no control of grammatical structures, with frequent errors that significantly obscure meaning

Score 1: 莫玩篮球房间。爸爸说莫它的。莫怕篮球房间。爸爸不高心有莫。爸爸莫去房

Score 1: 小明在床上。

Score of 2: Weak: Suggests lack of competence in interpersonal writing - TASK COMPLETION: 1) E-mail addresses topic only marginally or addresses only some aspects of stimulus 2) Scattered information generally lacks organization and coherence; minimal or no use of transitional elements and cohesive devices; fragmented sentences -DELIVERY: 1) Frequent use of register inappropriate to situation -LANGUAGE USE: 1) Minimal appropriate vocabulary, with frequent errors that obscure meaning; repeated interference from another language 2) Limited grammatical structures, with frequent errors that obscure meaning

Score 2: 小蔡在床上，爸爸来房间，爸爸回到房间看，儿子没有整理。

Score 2: 今天爸爸来到我的房间，让我打扫，可是我不想打扫。

Score of 3: Adequate Suggests competence in interpersonal writing -TASK COMPLETION: 1) E-mail addresses topic directly but may not address all aspects of stimulus 2) Portions may lack organization or coherence; infrequent use of transitional elements and cohesive devices; disconnected sentences -DELIVERY: 1) Use of register appropriate to situation is inconsistent or includes many errors -LANGUAGE USE: 1) Limited appropriate vocabulary and idioms, with frequent errors that sometimes obscure meaning; intermittent interference from another language 2) Mostly simple grammatical structures, with frequent errors that sometimes obscure meaning

Score 3: 一天在星期五一个人叫麦可，麦可的爸爸要告诉他如果你要玩在电视你要干净你的地上。然后麦可要干净你的地上，他的爸爸要去看还有看，他方他的篮球，衣服，皮扎在他的睡觉地方。他的爸爸很生气！他去给他讲话。爸爸说"我很生气在你，你不可以玩电视"！然后他说"去放你的篮球，衣服，和皮扎放起来"！

Score 3: 小李的房间太凌乱了。 小李爸爸进来他的房间，看到房间里的混乱。小李的爸爸告诉他 "收拾你的房间！不然我就要拿走你的手机了"。 所以小李床下面脏器了混乱。小李收拾完的 时候，他去看电影。他爸爸找到小李做的事，他生气了。他说小李得再收拾他的房间。

Score of 4: Good Demonstrates competence in interpersonal writing -TASK COMPLETION: 1) E-mail addresses all aspects of stimulus but may lack detail or elaboration 2) Generally organized and coherent; use of transitional elements and cohesive devices may be inconsistent; discourse of paragraph length, although sentences may be loosely connected -DELIVERY: 1) May include several lapses in otherwise consistent use of register appropriate to situation -LANGUAGE USE: 1) Mostly appropriate vocabulary and idioms, with errors that do not generally obscure meaning 2) Mostly appropriate grammatical structures, with errors that do not generally obscure meaning

Score 4: 儿子正在床上放松，他的爸爸走进他的房间，责骂他的居住环境凌乱不堪。爸爸要求儿子立刻收拾房间。然而，儿子没有打扫房间，而是把一切都塞到了床底下。爸爸后来回来，得知儿子没有打扫房间，而是把一切都藏在了床底下。爸爸下楼并要求儿子返回并打扫他的房间。

Score 4: 一个下午，小王在翻松看电视。可是，他爸爸走进了他的房间，告诉他"你的房间真乱啊！以前你可以看电视，你现在得打扫。" 所以小王把他的东西都放在床底下了，下楼。可是他的爸爸又走进了他的房间， 看床底下看到所有的物品，他下楼告诉小王"你还得打扫！"

Score of 5: Very good Suggests excellence in interpersonal writing -TASK COMPLETION: 1) E-mail addresses all aspects of stimulus 2) Well organized and coherent, with a progression of ideas that is generally clear; some use of transitional elements and cohesive devices; connected discourse of paragraph length -DELIVERY: 1) Consistent use of register appropriate to situation except for occasional lapses -LANGUAGE USE: 1) Appropriate vocabulary and idioms, with sporadic errors 2) Variety of grammatical structures, with sporadic errors

Score 5: 有一天下午，天明在房间里面听手机上的音乐。突然，天明的爸爸走进他的房间里，发现了他的房间乱七八糟，就对天明说他需要先把房间收拾好，才能玩儿。爸爸走了以后，天明就偷偷的把房间里面的所有东西都放在床底下。过了几个小时以后，爸爸又来到天明的房间里检查收拾了怎么样。天有不测风云，爸爸发现了床底下的东西，很吃惊。那时候，天明在楼下玩游戏。爸爸赶快到楼下很生气的跟天明说他需要把东西收拾好。

Score 5: 有一天，一个叫小明的男孩独自在房间里听音乐。他已经有几周还没有打扫了，房间很乱。当小明的爸爸进他的房间看到这个混乱时，他非常生气，立刻命令小明打扫。小明真的不想打扫房间，他宁愿看电视。爸爸离开的时候，他就制定计划。小明把所有的垃圾和脏衣服都收拾起来，然后藏在床底下。他以为这足以让他爸爸满意，赶快下到客厅看电视。起初，当小明的爸爸看到房间有多干净时，他感到自豪，但当他看床底下时，一切都改变了。他立刻下到客厅，告诉小明要真正打扫，不要只是把东西藏在床底下。

Score of 6: Excellent Demonstrates excellence in interpersonal writing -TASK COMPLETION: 1) E-mail addresses all aspects of stimulus with thoroughness and detail 2) Well organized and coherent, with a clear progression of ideas; use of appropriate transitional elements and cohesive devices; well-connected discourse of paragraph length -DELIVERY: 1) Consistent use of register appropriate to situation -LANGUAGE USE: 1) Rich and appropriate vocabulary and idioms, with minimal errors 2) Wide range of grammatical structures, with minimal errors

Score 6: 一个阳光明媚的下午，小明躺在床上，一边听音乐一边玩手机，悠闲得很。突然，爸爸推门而入，看到房间乱得不可收拾，顿时火冒三丈。地上散落着书本、袜子、鞋子、球类，甚至还有披萨盒。他厉声命令小明立刻整理房间。小明口头答应，可等爸爸一离开，他便把所有东西胡乱塞进床底，装作已经收拾好了。自以为聪明的小明满意地走下楼，准备看电影放松一下。爸爸起了疑心，觉得五分钟不可能彻底整理完，于是折返回来查看。房间表面看起来干净整齐，但他无意间发现床底露出一本书的角。他蹲下身，一掀床单，果然发现所有杂物都藏在下面。他无奈地摇头，大喊："小明，回房间来！"正在客厅吃东西的小明听见爸爸的声音，吓得立刻冲回楼上。

Score 6: 今天爸爸走进我的房间，看到地板上堆满了书本、衣服和零食包装袋，脸色立刻沉了下来。他盯着我，说："你必须把房间打扫干净。"我心里虽然知道他是对的，但实在太懒了，就随便把所有东西都塞到了床底下，假装收拾过了。然后我慢悠悠地下楼，一边吃着薯片一边看电视，脑子有点恍惚，完全没把这件事放在心上。没过多久，爸爸上楼检查，果然发现床底藏着所有的"秘密"。我的床底堆满了东西。他又下楼来，脸色更难看了，对我说："你必须重新认真整理房间。"我这才意识到，想蒙混过关是行不通的，我总好快速跑上楼去打扫。

Appendix B

Table B.1
Standard Deviation of the Holistic and Analytic Scores Assigned by Each Rater

|  | R1 | R2 | A1 | A2 | A3 | A4 | A5 | A6 | A7 | A8 | A9 | A10 |
|---|---|---|---|---|---|---|---|---|---|---|---|---|
| SN1 | 0.75 | 0.76 | 1.01 | 0.89 | 1.17 | 0.79 | 1.10 | 0.83 | 0.67 | 0.84 | 0.92 | 1.17 |
| ER1 | 0.88 | 0.95 | 1.47 | 0.68 | 0.91 | 0.90 | 1.22 | 0.76 | 0.85 | 1.24 | 1.30 | 1.19 |
| SN2 | 0.68 | 0.65 | 0.94 | 1.03 | 0.84 | 0.66 | 1.06 | 0.94 | 1.02 | 0.90 | 0.86 | 1.19 |
| ER2 | 1.11 | 1.17 | 1.01 | 1.23 | 1.25 | 0.67 | 1.38 | 1.03 | 1.06 | 1.17 | 1.47 | 1.46 |
| Task_SN1 | 1.03 | 0.85 | 0.79 | 0.79 | 1.25 | 0.79 | 1.10 | 0.90 | 0.68 | 1.15 | 1.02 | 0.84 |
| Task_SN2 | 0.89 | 0.78 | 0.82 | 0.92 | 1.07 | 0.66 | 1.03 | 1.16 | 0.74 | 1.04 | 0.68 | 1.11 |
| Task_ER1 | 1.20 | 1.06 | 1.53 | 0.82 | 1.18 | 1.13 | 1.30 | 0.83 | 0.86 | 1.09 | 1.31 | 1.40 |
| Task_ER2 | 1.40 | 1.28 | 1.17 | 1.23 | 1.25 | 0.73 | 1.28 | 1.00 | 0.98 | 1.10 | 1.44 | 1.46 |
| Delivery_SN1 | 0.86 | 0.85 | 1.07 | 0.89 | 1.14 | 0.88 | 1.10 | 0.80 | 0.65 | 0.93 | 0.92 | 1.17 |
| Delivery_SN2 | 0.63 | 0.67 | 0.86 | 0.92 | 0.87 | 0.66 | 1.05 | 0.68 | 0.72 | 0.90 | 0.86 | 1.05 |
| Delivery_ER1 | 0.86 | 0.98 | 1.39 | 0.90 | 0.89 | 0.77 | 1.16 | 0.76 | 0.77 | 0.93 | 1.20 | 1.17 |
| Delivery_ER2 | 1.35 | 1.28 | 1.03 | 1.07 | 1.22 | 0.81 | 1.21 | 1.00 | 0.81 | 0.97 | 1.04 | 1.28 |
| LUSE_SN1 | 0.87 | 0.93 | 1.04 | 0.89 | 1.03 | 0.97 | 1.10 | 0.76 | 0.77 | 1.12 | 0.98 | 1.01 |
| LUSE_SN2 | 0.60 | 0.68 | 0.86 | 0.97 | 0.81 | 0.66 | 1.03 | 0.76 | 1.24 | 0.82 | 0.78 | 1.09 |
| LUSE_ER1 | 0.77 | 0.89 | 1.40 | 0.72 | 1.03 | 0.93 | 1.01 | 0.56 | 0.89 | 0.89 | 1.16 | 1.01 |
| LUSE_ER2 | 1.30 | 1.30 | 0.92 | 1.24 | 1.01 | 0.78 | 1.22 | 1.00 | 0.91 | 1.00 | 0.91 | 1.19 |

Note: A1 – ChatGPT 3.5, A2 – ChatGPT 4, A3 – ChatGPT 4o, A4 - Gemini 1.5, A5 - Claude 3.5 Sonnet, A6 - OpenAI o1, A7 – Gemini 2.0, A8 – DeepSeek V3, A9 - Gemini 1.5 Pro, A10 - DeepSeek R1.